\documentclass{article}

\usepackage[T1]{fontenc}
\usepackage[utf8]{inputenc}

\PassOptionsToPackage{table,xcdraw}{xcolor}
\usepackage{hyperref}
\usepackage[preprint]{icml2026}
\makeatletter
\renewcommand{\Notice@String}{}
\makeatother

\usepackage{microtype}
\usepackage{graphicx}

\usepackage{amsmath,amssymb,amsfonts}
\usepackage{textcomp}
\usepackage{multirow}
\usepackage{hhline}

\providecommand{\changed}[1]{#1}

\icmltitlerunning{MoP: Mixture of Pruners}

\begin{document}

\twocolumn[
  \icmltitle{Compressing LLMs with MoP:\texorpdfstring{\\}{ }%
    \texorpdfstring{\underline{M}}{M}ixture
    \texorpdfstring{\underline{o}}{o}f
    \texorpdfstring{\underline{P}}{P}runers}

  \begin{icmlauthorlist}
    \icmlauthor{Bruno Lopes Yamamoto}{usp}
    \icmlauthor{Lucas Lauton de Alcantara}{usp}
    \icmlauthor{Victor Zacarias}{usp}
    \icmlauthor{Leandro Giusti Mugnaini}{usp}
    \icmlauthor{Keith Ando Ogawa}{usp}
    \icmlauthor{Lucas Pellicer}{icti}
    \icmlauthor{Rosimeire Pereira Costa}{icti}
    \icmlauthor{Edson Bollis}{icti}
    \icmlauthor{Anna Helena Reali Costa}{usp}
    \icmlauthor{Artur Jordao}{usp}
  \end{icmlauthorlist}

  \icmlaffiliation{usp}{Universidade de São Paulo, Brazil}
  \icmlaffiliation{icti}{Instituto de Ciência e Tecnologia Itaú (ICTi), Brazil}

  \icmlcorrespondingauthor{}{brunolyamamoto@usp.br}

  \icmlkeywords{LLM, Pruning, Multimodal, Efficient Models}

  \vskip 0.3in
]

\printAffiliationsAndNotice{}

\begin{abstract}

The high computational demands of Large Language Models (LLMs) motivate methods that reduce parameter count and accelerate inference. In response, model pruning emerges as an effective strategy, yet current methods typically focus on a single dimension—depth or width. We introduce MoP (Mixture of Pruners), an iterative framework that unifies these dimensions. At each iteration, MoP generates two branches—pruning in depth versus pruning in width—and selects a candidate to advance the path. On LLaMA-2 and LLaMA-3, MoP advances the frontier of structured pruning, exceeding the accuracy of competing methods across a broad set of compression regimes. It also consistently outperforms depth-only and width-only pruning. Furthermore, MoP translates structural pruning into real speedup, reducing end-to-end latency by 39\% at 40\% compression. Finally, extending MoP to the vision-language model LLaVA-1.5, we notably improve computational efficiency and demonstrate that text-only recovery fine-tuning can restore performance even on visual tasks.
\end{abstract}

\begin{figure}[t]
    \centering
    \includegraphics[width=0.9\linewidth]{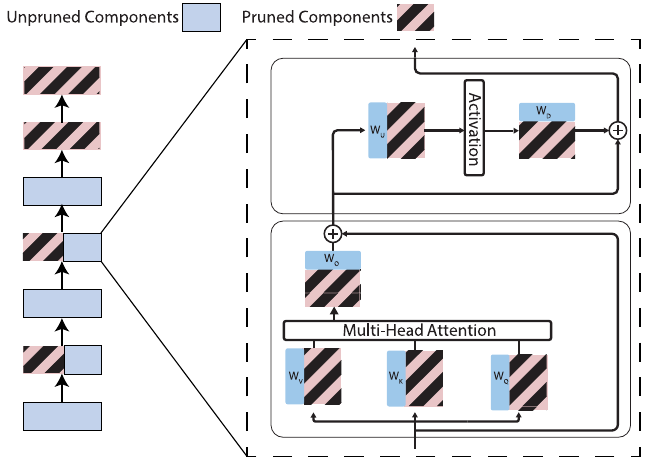}
    \caption{Illustration of MoP combining depth pruning (removing transformer layers) and width pruning (removing attention heads and MLP neurons).}
    \label{fig:teaser}
\end{figure}

\section{Introduction}\label{sec:introduction}

\changed{In the rapidly transforming world of computation, large foundational models stand as the primary drivers of cognitive progress, pushing the frontier of machine intelligence to a previously unseen degree~\cite{Maslej:2025}}. Such outstanding performance often entails unintended downsides, especially as models scale in size and cost~\cite{Maslej:2025}. In this context, model compression serves as an effective means to reduce model size and, in turn, computational demands~\cite{Ma:2023,Sreenivas:2024}. A practical route to this goal is pruning, particularly structured pruning.

Structured pruning removes coherent structures rather than individual weights~\cite{Ma:2023}. Unlike unstructured sparsity, these methods do not require specialized hardware or software to realize the speedup and cost reduction~\cite{Ashkboos:2024}. Within structured pruning, two main approaches emerge. The first removes stacked components (e.g., entire transformer layers) to reduce \changed{model depth}, resulting in a shallower architecture (depth pruning). The second removes components within layers (e.g., attention heads, MLP neurons) to reduce internal dimensions, resulting in a narrower architecture (width pruning). While width pruning enables finer-grained selection of redundant structures, tending to better preserve \changed{performance on downstream} tasks~\cite{Sreenivas:2024}, depth pruning, \changed{by reducing sequential computation}, often yields superior inference acceleration~\cite{Sreenivas:2024}.

\begin{figure}[!bt]
  \centering
  \includegraphics[width=0.9\linewidth]{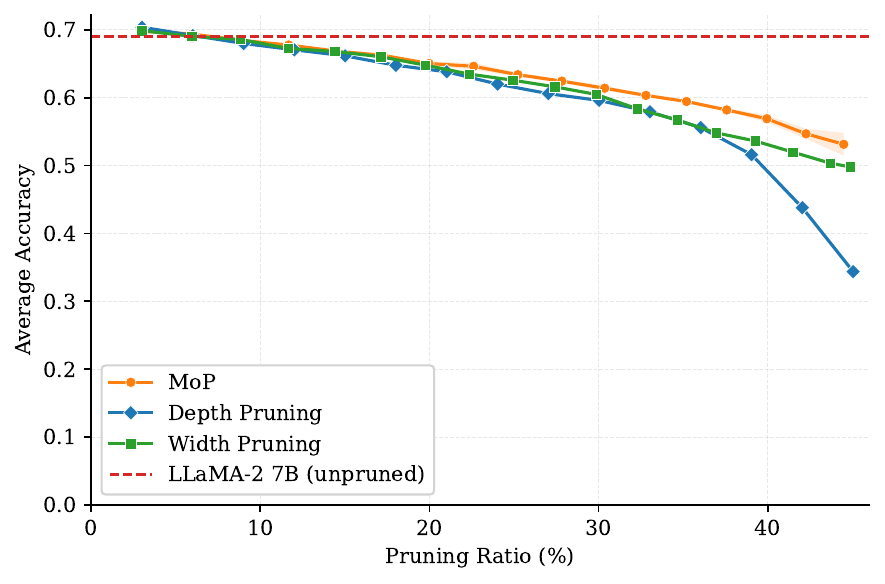}
  \caption{Average accuracy on \changed{five standard commonsense} benchmarks, comparing depth pruning, width pruning, and MoP (combining both) on LLaMA-2 7B. MoP achieves a Pareto improvement over single-dimension approaches.}
  \label{fig:mop_vs_extremes}
\end{figure}

\changed{Building on these complementary strengths, we introduce our Mixture of Pruners (MoP) method, a novel approach to the depth--width dichotomy that synthesizes both benefits through an iterative pruning process. Figure~\ref{fig:mop_vs_extremes} summarizes the advantages of our strategy and confirms that} MoP consistently outperforms both pure depth and pure width pruning across all compression ratios on LLaMA-2 7B, achieving a Pareto improvement over single-dimension approaches.

MoP builds upon existing \changed{forms of pruning} through an iterative process. At each step, our method evaluates both depth and width candidates, using a criterion to guide the selection. This process repeats until reaching the target compression, followed by recovery fine-tuning. \changed{The modular design accommodates different pruning criteria. We further demonstrate that MoP extends to multimodal architectures such as LLaVA~\cite{llava2024improved}, and we observe that text-only recovery fine-tuning can substantially restore multimodal performance after pruning (see Section~\ref{sec:mop-multimodal}).}

 Among our contributions, we highlight the following: (1) We propose MoP, an efficient method that effectively combines both depth and width pruning, capturing the advantages of both paradigms: the latency reduction of depth pruning and the fine-grained selectivity of width pruning.
(2) MoP follows a modular design, meaning it accommodates other combinations of width and depth pruning criteria. \changed{This allows MoP to incorporate future state-of-the-art pruning criteria, possibly further enhancing its performance.}
(3) MoP provides a more diverse pruning scheme, achieving higher compression ratios by distributing removal across both depth and width rather than exhausting a single dimension. 
(4) Unlike most pruning literature, we test our method in both language and multimodal architectures.
\changed{(5) To the best of our knowledge, we are the first to observe that text-only recovery fine-tuning can, surprisingly, restore performance on visual benchmarks in pruned vision-language models.}

\begin{sloppypar}
Code and models are available at: \url{https://github.com/c2d-usp/Efficient-LLMs-with-MoP}.
\end{sloppypar}

\section{Related Work}\label{sec:related}

\noindent \textbf{Depth pruning.} Within a layer-wise pruning approach, LLM-Streamline~\cite{Chen:2025} measures the importance of layers by assessing the cosine similarity between the input and output hidden states of each of them. The authors argue that a high cosine similarity indicates a nonessential transformation, suggesting that the layer is a strong candidate for pruning. In a different direction, LaCo~\cite{Yang:2025} merges layers to achieve model compression. Specifically, the process iteratively merges a set of contiguous layers, assesses whether the merged layer is similar enough to the original set, and selects the next set of layers to merge. 
To better understand how LLMs store knowledge in their weights, Gromov et al.~\cite{Gromov:2025} leverage cosine similarity to analyze layer pruning. Interestingly, they found that, for most architectures, promising pruning sets of layers appear in the deeper part of the neural network. Based on this evidence, the authors show that removing the deepest layers leads to a model with predictive performance on question-answering tasks comparable to carefully selecting with a similarity measure. Building on these findings, we follow a similar strategy for the layer pruning part of our method MoP.

\noindent \textbf{Width pruning. } Aiming at attention heads and MLP neurons, AMP~\cite{Mugnaini:2025} measures the importance of these components by assessing the magnitude of their activation. The authors associate a lower magnitude of activation with a less important contribution and uniformly prune across all layers. On the other hand, PruneNet~\cite{YOPO:2025} focuses on pruning MLPs by looking at the singular value distribution of their weight matrices. Rather than using data to assess important structures, the authors propose a policy-based technique that learns the rows to prune. \changed{ModHiFi~\cite{modhifi2025} introduces Subset Fidelity, a local reconstruction metric with an optimal compensation term to score component importance without gradients. ModHiFi then keeps high-fidelity channels based on scores and prunes the rest using synthetic calibration data.}

In contrast to previous work, MoP benefits from characteristics of both types of structured pruning strategies. More concretely, our method leverages the superior inference acceleration of layer-wise pruning and the finer granularity of component-wise pruning. We also extend our method to visual-language architectures.

\section{Proposed Mixture of Pruners (MoP) method}
\noindent
\subsection{Transformer Architecture}
\noindent
\textbf{Depth Aspects.} Most modern language transformers employ a decoder-only architecture~\cite{grattafiori2024llama,Touvron:2023,Touvron2:2023,Li:2023}, essentially containing a single deep stack of layers that consecutively transform the input into a meaningful final representation.
Therefore, model depth depends on the number of stacked layers. Each layer shares the same general structure: an attention module and a multi-layer perceptron (MLP), with normalization and skip connections. Importantly, all layers share the same dimensions. This leads to two key consequences for depth pruning. First, depth-wise pruning removes layers without changing the model architecture or adding extra parameters, since the inputs and outputs of every layer are dimensionally compatible. Second, the full layer is the smallest removable depth unit that still preserves model structure, since removing internal components (e.g., the MLP or attention module) would make the architecture non-uniform across layers.

\noindent
\textbf{Width Aspects.} In a transformer layer, most parameters belong to the attention module and the MLP. Naturally, these modules are the usual target for pruning methods within a layer~\cite{Mugnaini:2025,YOPO:2025}. In the attention and MLP modules, all learnable parameters reside in the linear projection matrices. Width pruning, therefore, consists of reducing the internal dimensions of these matrices by removing whole units (e.g., attention heads or MLP neurons), thereby reducing their size. In most cases, width pruning narrows only the internal widths, thus keeping input and output compatibility between layers. Since width pruning works on internal dimensions, which are typically in the thousands, it allows much finer granularity, for example, acting at the level of individual neurons in the MLP. 
Overall, depth and width pruning methods operate on different axes: the former reduces the transformer by decreasing the number of stacked layers, whereas the latter decreases the dimensions of the wide projection matrices inside each layer to obtain a narrower structure. Figure~\ref{fig:teaser} illustrates this distinction.

\begin{algorithm}[!h]
    \small
    \caption{\changed{Pruning with MoP}}
    \label{alg::mop}
    \begin{algorithmic}[1]
        \item[] \textbf{Input:} Unpruned model $\mathcal{F}$, Target compression ratio $\rho$, Training samples $\mathcal{D}$ with labels $\mathbf{Y}$, Fixed calibration set $\mathcal{D}_{\text{cal}}$, Path criterion $\mathcal{P}$, Width pruning criterion $\mathcal{W}$, Layer pruning criterion $\mathcal{L}$
        \item[] \textbf{Output:} $\mathcal{F}'$ (Pruned version of $\mathcal{F}$)\\

        \STATE $\mathcal{F}^{(0)} \gets \mathcal{F}$  \hfill $\triangleright$ Initialise current model
        \STATE $t \gets 1$
        \STATE $P_{\min} \gets (1 - \rho) \cdot \textsc{TotalParams}(\mathcal{F})$ \hfill $\triangleright$ Target parameter count
        
        \WHILE{$\textsc{TotalParams}(\mathcal{F}^{(t-1)}) > P_{\min}$}
            \STATE $\mathcal{D}^{(t)}_{\text{tune}} \gets \textsc{SampleTuneSubset}(\mathcal{D})$ \hfill $\triangleright$ Iteration-dependent fine-tuning subset
            \STATE $\ell^\star \gets \mathcal{L}\!\bigl(\mathcal{F}^{(t-1)}\bigr)$ \hfill $\triangleright$ Select one layer to remove
            \STATE $p_{\ell} \gets \textsc{LayerParams}\!\bigl(\mathcal{F}^{(t-1)}, \ell^\star\bigr)$ \hfill $\triangleright$ Parameter count of the selected layer
            \STATE $c_t \gets p_{\ell} \,/\, \textsc{TotalParams}\!\bigl(\mathcal{F}^{(t-1)}\bigr)$ \hfill $\triangleright$ Compression ratio based on that layer
            
            \STATE $\triangleright$ \textbf{Width-level candidate}
            \STATE $\Omega \gets \mathcal{W}\!\bigl(\mathcal{F}^{(t-1)}, \mathcal{D}_{\text{cal}}\bigr)$ \hfill $\triangleright$ Score width units by $\mathcal{W}$ on calibration data
            \STATE $\mathcal{F}_{\text{width}} \gets \textsc{WidthPrune}\!\bigl(\mathcal{F}^{(t-1)}, \Omega, c_t\bigr)$
            \STATE $\tilde{\mathcal{F}}_{\text{width}} \gets \textsc{FineTune}\!\bigl(\mathcal{F}_{\text{width}}, \mathcal{D}^{(t)}_{\text{tune}}\bigr)$
            \STATE $s_{\text{width}} \gets \mathcal{P}\!\bigl(\tilde{\mathcal{F}}_{\text{width}}, \mathcal{F}, \mathcal{D}_{\text{cal}}\bigr)$ \hfill $\triangleright$ Compare against original model
            
            \STATE $\triangleright$ \textbf{Layer-level candidate}
            \STATE $\mathcal{F}_{\text{layer}} \gets \textsc{LayerPrune}\!\bigl(\mathcal{F}^{(t-1)}, \ell^\star\bigr)$
            \STATE $\tilde{\mathcal{F}}_{\text{layer}} \gets \textsc{FineTune}\!\bigl(\mathcal{F}_{\text{layer}}, \mathcal{D}^{(t)}_{\text{tune}}\bigr)$
            \STATE $s_{\text{layer}} \gets \mathcal{P}\!\bigl(\tilde{\mathcal{F}}_{\text{layer}}, \mathcal{F}, \mathcal{D}_{\text{cal}}\bigr)$ \hfill $\triangleright$ Compare against original model
            
            \STATE $\triangleright$ \textbf{Path decision}
            \IF{$s_{\text{width}} \le s_{\text{layer}}$}
                \STATE $\mathcal{F}^{(t)} \gets \mathcal{F}_{\text{width}}$
            \ELSE
                \STATE $\mathcal{F}^{(t)} \gets \mathcal{F}_{\text{layer}}$
            \ENDIF
            \STATE $t \gets t + 1$
        \ENDWHILE
        
        \STATE $\mathcal{F}' \gets \textsc{FineTune}\!\bigl(\mathcal{F}^{(t-1)}, \mathcal{D}, \mathbf{Y}\bigr)$ 
    \end{algorithmic}
\end{algorithm}

\subsection{Proposed method}

\changed{For clarity, we first describe the general MoP framework under generic width and depth pruning criteria, and then instantiate it with the specific choices adopted in this work.}

\noindent\textbf{Width and Depth pruning.}
For the depth-wise pruning, we choose to remove at each step one complete transformer layer. This choice provides the minimal pruning unit along the depth dimension, since removing submodules inside a transformer layer (e.g., entire linear projections in the attention or MLP) would either make internal dimensions incompatible or remove essential transformer structures.

In the case of width pruning, MoP works with any width pruning method that provides enough granularity in the pruning ratio (i.e., small steps). This contrasts with the coarse jumps in model size that layer removal produces. As we shall see, this condition is necessary so that we can match the number of parameters pruned when removing a layer or pruning in width.

\noindent
\textbf{MoP algorithm.} We build our method upon the idea of exploring the advantages of pruning structures at different granularities \changed{(i.e., layers and their width components)}, challenging the current dichotomy of depth and width pruning. Instead of removing only one type of component, we propose to iteratively evaluate both based on existing pruning criteria. This modular approach allows us to integrate new pruning strategies for width or depth into our method.
Formally, given a model $\mathcal{F}$ and a pair of pruning methods, one for width and one for depth, to achieve a desired compression ratio, we iteratively prune the model. At each iteration, a path criterion $\mathcal{P}$ decides which type of pruning to apply, resulting in a final model that combines both strategies.

A question that arises is how much to prune at each iteration. We want the smallest step possible to maintain fine control over the pruning trajectory. The minimum for layer pruning is one layer, so we use this as the basis and prune the equivalent number of parameters in width, \changed{this} ensures a fair comparison between both candidates. Since width pruning is more fine-grained, it can approximate the number of parameters of one layer. At each iteration, we first identify the layer $\ell^\star$ selected by criterion $\mathcal{L}$, then compute $c_t$ as the fraction of parameters that the selected layer represents (line 8 in Algorithm~\ref{alg::mop}). Importantly, we recompute $c_t$ at each iteration since width pruning changes the number of parameters per layer. This ensures that we compare both candidates on a consistent basis, removing the same number of parameters at each iteration.

Algorithm~\ref{alg::mop} details our method. At each iteration $t$, we create two candidate models: one by applying width pruning with ratio $c_t$ and one by removing a single layer. The width criterion $\mathcal{W}$ produces the width candidate $\mathcal{F}_{\text{width}}$, and the layer criterion $\mathcal{L}$ selects which layer to remove, producing $\mathcal{F}_{\text{layer}}$. We then apply a short recovery fine-tuning step to both candidates on a training subset $\mathcal{D}^{(t)}_{\text{tune}}$ before evaluation. The path criterion $\mathcal{P}$ scores the corresponding fine-tuned candidates $\tilde{\mathcal{F}}_{\text{width}}$ and $\tilde{\mathcal{F}}_{\text{layer}}$ against the original (unpruned) model and decides whether width or depth pruning defines the next model $\mathcal{F}^{(t)}$. We set $\mathcal{F}^{(t)}$ to the corresponding pruned model \emph{before} this short update, $\mathcal{F}_{\text{width}}$ or $\mathcal{F}_{\text{layer}}$, discarding the intermediate fine-tuning, \changed{serving only to assess the relative quality of the candidates.} The process repeats until we reach the desired compression, and we then apply full fine-tuning to the final model on the complete training set $\mathcal{D}$.

\noindent
\textbf{The Layer criterion.}
Recent literature on layer pruning explores data-driven methods for ascertaining redundancy in transformer layers~\cite{Gromov:2025,Chen:2025}. These methods, based on criteria such as cosine similarity, converge to a common finding: the deeper layers in the model tend to be more redundant, with the exception of the very last, which is essential due to its proximity to the output projection. According to Gromov et al.~\cite{Gromov:2025}, their data-driven approach actually leads to a simple algorithm: prune from the penultimate layer to the first. Kim et al.~\cite{Kim:2024} use a slightly higher margin and preserve the last two layers instead of just one. Following these insights from the literature, we adopt as our layer criterion $\mathcal{L}$ the following procedure. \changed{At each iteration, the criterion $\mathcal{L}$ selects the third-to-last layer for pruning.} This guarantees that we always keep the last two layers and that we remove layers from the end to the beginning.

\noindent    
\textbf{The Width Criterion.} Adhering to the requirements of fine-grained compression control and strong empirical performance, we adopt AMP~\cite{Mugnaini:2025} as the width pruning criterion $\mathcal{W}$. AMP removes attention heads and MLP neurons uniformly across layers. This neuron-level pruning enables MoP to precisely match the parameter count of a single layer at each iteration, ensuring fair comparisons between width and depth candidates. Additionally, AMP outperforms several width pruning methods on benchmarks for the LLaMA family~\cite{Mugnaini:2025}.

\noindent
\textbf{The Path Criterion.} At each iteration, MoP must decide whether to apply width or depth pruning. A path criterion $\mathcal{P}$ governs this decision and evaluates both candidate models ($\tilde{\mathcal{F}}_{\text{width}}$ and $\tilde{\mathcal{F}}_{\text{layer}}$) and selects the one that best preserves the original model's capabilities. We define $\mathcal{P}$ to output a single scalar score that measures how much a candidate deviates from the original (unpruned) model, where smaller scores indicate closer behavior and are therefore preferred. To ensure a fair comparison between candidates, MoP prunes both with the same compression ratio $c_t$, removing an equivalent number of parameters regardless of the pruning type.
    
MoP is modular in its choice of path criterion $\mathcal{P}$. In this work, we instantiate $\mathcal{P}$ \changed{with} commonly used metrics\changed{~\cite{agarwal2024onpolicy,muralidharan2024compact,Gromov:2025,Kim:2024,Mugnaini:2025,Chen:2025}}, including cosine similarity, KL divergence, and perplexity, and we also test randomized path selection.

\subsection{Path Selection Strategies}
\label{exp:path}

\changed{
 We explore three distinct metrics for $\mathcal{P}$: cosine similarity, Kullback–Leibler (KL) divergence, and perplexity (PPL). These rely on a simple forward pass with the calibration data subset. In all cases, lower values indicate closer agreement with the reference model. We also test random selection as a baseline.
 
\noindent
\textbf{Cosine similarity.} We compute the angle~\cite{Gromov:2025} between the flattened output logits of the reference and pruned models, $\mathbf{v}_{ref}$ and $\mathbf{v}_{pruned}$:
\begin{equation}
    \theta = \arccos\left( \frac{\mathbf{v}_{ref} \cdot \mathbf{v}_{pruned}}{\|\mathbf{v}_{ref}\|_2 \|\mathbf{v}_{pruned}\|_2} \right).
\end{equation}

\noindent
\textbf{Kullback-Leibler divergence.} We adopt Kullback-Leibler~\cite{agarwal2024onpolicy,muralidharan2024compact} divergence to measure the distributional shift between the output token probabilities of the reference (unpruned) and pruned models. Let $p$ and $q$ denote the corresponding softmax distributions over the vocabulary $\mathcal{V}$. We compute the average divergence across $N$ sequence positions:
\begin{equation}
    D_{KL}(p \| q) = \frac{1}{N} \sum_{i=1}^{N} \sum_{x \in \mathcal{V}} p_i(x) \log \left( \frac{p_i(x)}{q_i(x)} \right).
\end{equation}

\noindent
\textbf{Perplexity.} We compute perplexity of the candidate model on calibration data~\cite{Gromov:2025,Kim:2024,Mugnaini:2025}, measuring how well the compressed model predicts a sequence of tokens $X = (x_1, \dots, x_T)$:
\begin{equation}
    \text{PPL}(X) = \exp \left( - \frac{1}{T} \sum_{t=1}^{T} \log P(x_t \mid x_1, \dots, x_{t-1}) \right).
\end{equation}

\noindent
\textbf{Randomized Path Selection.}
\label{exp:random}
As an alternative to metric-based selection, we also evaluate a randomized criterion that chooses between width and depth pruning with equal probability at each iteration.

}

\section{Experiments}\label{sec:experiments}

\begin{table}[!tb]
\centering
\caption{Performance comparison of path selection criteria for MoP on LLaMA 7B at a 30\% compression ratio. We highlight the highest accuracy across criteria in bold.}
\smallskip
\renewcommand{\arraystretch}{1.2}
\resizebox{\columnwidth}{!}{%
\begin{tabular}{c|ccccc|c}
\hline
Method         & WinoGrande    & HellaSwag    & ARC-e        & ARC-c        & PIQA         & Avg            \\ \hline
KL             & \textbf{65.27}         & 63.42        & 58.50        & 37.88        & 71.00        & 59.21          \\
\rowcolor[HTML]{EFEFEF}
Cosine         & 62.19         & \textbf{67.58}        & 62.71        & 37.46        & 73.45        & 60.68          \\
PPL            & 63.69         & 66.35        & 62.54        & \textbf{38.14}        & 73.50        & \textbf{60.84} \\
\rowcolor[HTML]{EFEFEF}
Random         & 63.14$\pm$0.91 & 66.47$\pm$0.38 & \textbf{63.01$\pm$0.52} & 37.71$\pm$0.60 & \textbf{73.81$\pm$0.60} & 60.83$\pm$0.43 \\ \hline
\end{tabular}
}
\label{tab:ablation}
\end{table}

\begin{table*}[!tb]
\centering
\caption{Performance of MoP relative to other \changed{state-of-the-art} pruning methods. We highlight the best results in bold for each compression ratio and model.}
\resizebox{\textwidth}{!}{%
\begin{tabular}{c|c|ccccc|c}
\hline
Pruning Ratio & Method & WinoGrande & HellaSwag & ARC-e & ARC-c & PIQA & Avg \\ \hhline{=|=|=====|=}

0\% & LLaMA-2 7B & 69.14 & 75.99 & 74.58 & 46.15 & 79.11 & 68.99 \\ \hline

& \cellcolor[HTML]{EFEFEF}SliceGPT~\cite{Ashkboos:2024} (ICLR, 2024) & \cellcolor[HTML]{EFEFEF}65.11 & \cellcolor[HTML]{EFEFEF}59.04 & \cellcolor[HTML]{EFEFEF}59.76 & \cellcolor[HTML]{EFEFEF}37.54 & \cellcolor[HTML]{EFEFEF}69.42 & \cellcolor[HTML]{EFEFEF}58.18 \\
& AmoebaLLM~\cite{fu2024amoeballm} (NeurIPS, 2024) & 66.90 & 70.80 & 70.30 & 40.20 & 76.30 & 64.90 \\
& \cellcolor[HTML]{EFEFEF}PruneNet~\cite{YOPO:2025} (ICLR, 2025) & \cellcolor[HTML]{EFEFEF}66.22 & \cellcolor[HTML]{EFEFEF}68.37 & \cellcolor[HTML]{EFEFEF}63.93 & \cellcolor[HTML]{EFEFEF}38.40 & \cellcolor[HTML]{EFEFEF}74.76 & \cellcolor[HTML]{EFEFEF}62.34 \\
& SlimLLM~\cite{guo2025slimllm} (ICML, 2025) & 64.88 & 70.95 & 67.17 & 38.99 & \textbf{78.02} & 64.00 \\
& \cellcolor[HTML]{EFEFEF}ModHiFi-P~\cite{modhifi2025} (NeurIPS, 2025) & \cellcolor[HTML]{EFEFEF}64.64 & \cellcolor[HTML]{EFEFEF}62.70 & \cellcolor[HTML]{EFEFEF}64.73 & \cellcolor[HTML]{EFEFEF}38.22 & \cellcolor[HTML]{EFEFEF}72.79 & \cellcolor[HTML]{EFEFEF}60.62 \\
& CoMe~\cite{Wang:2025CoMe} (NeurIPS, 2025) & 67.25 & 68.68 & 64.10 & 39.59 & 72.42 & 62.41 \\
& \cellcolor[HTML]{EFEFEF}LINEARPATCH~\cite{LinearPatch:2025} (NeurIPS, 2025) &
\cellcolor[HTML]{EFEFEF}\textbf{67.40} & \cellcolor[HTML]{EFEFEF}69.33 & \cellcolor[HTML]{EFEFEF}64.35 & \cellcolor[HTML]{EFEFEF}38.23 & \cellcolor[HTML]{EFEFEF}73.23 & \cellcolor[HTML]{EFEFEF}62.51 \\
& AMP~\cite{Mugnaini:2025} (IJCNN, 2025) & 61.56 & 69.22 & 68.18 & 42.06 & 76.39 & 63.48 \\
\multirow{-9}{*}{20\%} & \cellcolor[HTML]{EFEFEF}\textbf{MoP (Ours)} & \cellcolor[HTML]{EFEFEF}65.19$\pm$1.12 & 
\cellcolor[HTML]{EFEFEF}\textbf{71.16$\pm$0.51} & \cellcolor[HTML]{EFEFEF}\textbf{70.54$\pm$0.60} & \cellcolor[HTML]{EFEFEF}\textbf{42.86$\pm$0.32} & \cellcolor[HTML]{EFEFEF}76.35$\pm$0.06 & \cellcolor[HTML]{EFEFEF}\textbf{65.22$\pm$0.21} \\ \hline

& \cellcolor[HTML]{EFEFEF}SliceGPT~\cite{Ashkboos:2024} (ICLR, 2024) & \cellcolor[HTML]{EFEFEF}61.33 & \cellcolor[HTML]{EFEFEF}49.62 & \cellcolor[HTML]{EFEFEF}51.77 & \cellcolor[HTML]{EFEFEF}31.23 & \cellcolor[HTML]{EFEFEF}63.55 & \cellcolor[HTML]{EFEFEF}51.50 \\
& Shortened LLaMA~\cite{Kim:2024} (ICLR, 2024) & 61.09 & 54.97 & 45.96 & 34.81 & 60.99 & 51.56 \\
& \cellcolor[HTML]{EFEFEF}DISP-LLM~\cite{Gao:2024disp} (NeurIPS, 2024) & \cellcolor[HTML]{EFEFEF}63.93 & \cellcolor[HTML]{EFEFEF}62.87 & \cellcolor[HTML]{EFEFEF}60.10 & \cellcolor[HTML]{EFEFEF}37.03 & \cellcolor[HTML]{EFEFEF}73.72 & \cellcolor[HTML]{EFEFEF}59.53 \\
& PruneNet~\cite{YOPO:2025} (ICLR, 2025) & 62.90 & 63.21 & 53.37 & 33.70 & 72.20 & 57.08 \\
& \cellcolor[HTML]{EFEFEF}Yang et al.~\cite{Yang:2025Tell} (ICML, 2025) & \cellcolor[HTML]{EFEFEF}60.77 & \cellcolor[HTML]{EFEFEF}59.18 & \cellcolor[HTML]{EFEFEF}58.16 & \cellcolor[HTML]{EFEFEF}34.22 & \cellcolor[HTML]{EFEFEF}70.95 & \cellcolor[HTML]{EFEFEF}56.65 \\
& ModHiFi-P~\cite{modhifi2025} (NeurIPS, 2025) & 59.35 & 50.61 & 53.15 & 32.50 & 66.59 & 52.44 \\
& \cellcolor[HTML]{EFEFEF}CoMe~\cite{Wang:2025CoMe} (NeurIPS, 2025) &
\cellcolor[HTML]{EFEFEF}63.38 & \cellcolor[HTML]{EFEFEF}65.83 & \cellcolor[HTML]{EFEFEF}63.51 & \cellcolor[HTML]{EFEFEF}35.58 & \cellcolor[HTML]{EFEFEF}74.05 & \cellcolor[HTML]{EFEFEF}60.47 \\
& LINEARPATCH~\cite{LinearPatch:2025} (NeurIPS, 2025) &
\textbf{66.69} & 64.52 & 60.65 & 34.81 & 70.29 & 59.39 \\
& \cellcolor[HTML]{EFEFEF}AMP~\cite{Mugnaini:2025} (IJCNN, 2025) & \cellcolor[HTML]{EFEFEF}61.25 & \cellcolor[HTML]{EFEFEF}65.47 & \cellcolor[HTML]{EFEFEF}64.31 & \cellcolor[HTML]{EFEFEF}\textbf{39.85} & \cellcolor[HTML]{EFEFEF}\textbf{74.21} & \cellcolor[HTML]{EFEFEF}61.02 \\
\multirow{-10}{*}{30\%} & \textbf{MoP (Ours)} & 62.38$\pm$1.87 & \textbf{66.88$\pm$0.21} & \textbf{65.38$\pm$0.44} & 39.36$\pm$0.13 & 73.72$\pm$0.38 & \textbf{61.54$\pm$0.27} \\ \hline

& \cellcolor[HTML]{EFEFEF}Yang et al.~\cite{Yang:2025Tell} (ICML, 2025) & \cellcolor[HTML]{EFEFEF}56.67 & \cellcolor[HTML]{EFEFEF}49.00 & \cellcolor[HTML]{EFEFEF}49.83 & \cellcolor[HTML]{EFEFEF}30.38 & \cellcolor[HTML]{EFEFEF}66.10 & \cellcolor[HTML]{EFEFEF}50.39 \\
& AMP~\cite{Mugnaini:2025} (IJCNN, 2025) &
52.96 & 55.44 & 55.43 &
32.42 & \textbf{69.04} & 53.06 \\
\multirow{-3}{*}{40\%} & \cellcolor[HTML]{EFEFEF}\textbf{MoP (Ours)} &
\cellcolor[HTML]{EFEFEF}\textbf{59.35$\pm$1.32} & \cellcolor[HTML]{EFEFEF}\textbf{57.94$\pm$0.43} & \cellcolor[HTML]{EFEFEF}\textbf{57.05$\pm$1.20} & \cellcolor[HTML]{EFEFEF}\textbf{34.16$\pm$0.80} & \cellcolor[HTML]{EFEFEF}68.36$\pm$0.97 & \cellcolor[HTML]{EFEFEF}\textbf{55.37$\pm$0.45} \\ \hline \hline

0\% & LLaMA-3 8B & 72.77 & 79.19 & 77.78 & 52.90 & 80.85 & 72.70 \\ \hline

& \cellcolor[HTML]{EFEFEF}LINEARPATCH~\cite{LinearPatch:2025} (NeurIPS, 2025) & \cellcolor[HTML]{EFEFEF}70.17 & \cellcolor[HTML]{EFEFEF}66.74 & \cellcolor[HTML]{EFEFEF}60.82 & \cellcolor[HTML]{EFEFEF}43.17 & \cellcolor[HTML]{EFEFEF}72.85 & \cellcolor[HTML]{EFEFEF}62.75 \\
& CoMe~\cite{Wang:2025CoMe} (NeurIPS, 2025) & \textbf{70.96} & 65.52 & 64.23 & 40.44 & 73.50 & 62.93 \\
& \cellcolor[HTML]{EFEFEF}AMP~\cite{Mugnaini:2025} (IJCNN, 2025) &
\cellcolor[HTML]{EFEFEF}59.91 & \cellcolor[HTML]{EFEFEF}66.53 & \cellcolor[HTML]{EFEFEF}67.68 &
\cellcolor[HTML]{EFEFEF}44.20 & \cellcolor[HTML]{EFEFEF}75.30 & \cellcolor[HTML]{EFEFEF}62.72 \\
\multirow{-4}{*}{20\%} & \textbf{MoP (Ours)} & 66.80$\pm$1.34 & \textbf{71.94$\pm$0.13} & \textbf{67.94$\pm$1.45} & \textbf{44.62$\pm$0.77} & \textbf{75.88$\pm$1.80} & \textbf{65.44$\pm$1.04} \\ \hline

& \cellcolor[HTML]{EFEFEF}CoMe~\cite{Wang:2025CoMe} (NeurIPS, 2025) & \cellcolor[HTML]{EFEFEF}60.62 & \cellcolor[HTML]{EFEFEF}50.42 & \cellcolor[HTML]{EFEFEF}50.67 & \cellcolor[HTML]{EFEFEF}33.70 & \cellcolor[HTML]{EFEFEF}67.08 & \cellcolor[HTML]{EFEFEF}52.50 \\
& Yang et al.~\cite{Yang:2025Tell} (ICML, 2025) & 60.30 & 54.40 & 54.76 & 33.19 & 69.10 & 54.34 \\
& \cellcolor[HTML]{EFEFEF}AMP~\cite{Mugnaini:2025} (IJCNN, 2025) &
\cellcolor[HTML]{EFEFEF}60.77 & \cellcolor[HTML]{EFEFEF}61.93 & \cellcolor[HTML]{EFEFEF}59.76 &
\cellcolor[HTML]{EFEFEF}35.84 & \cellcolor[HTML]{EFEFEF}71.76 & \cellcolor[HTML]{EFEFEF}58.01 \\
\multirow{-4}{*}{30\%} & \textbf{MoP (Ours)} & \textbf{63.75$\pm$0.67} & \textbf{65.44$\pm$0.72} & \textbf{61.98$\pm$1.63} & \textbf{38.77$\pm$1.03} & \textbf{72.22$\pm$0.30} & \textbf{60.43$\pm$0.59} \\ \hline

\multirow{3}{*}{40\%}
& \cellcolor[HTML]{EFEFEF}Yang et al.~\cite{Yang:2025Tell} (ICML, 2025) & \cellcolor[HTML]{EFEFEF}56.27 & \cellcolor[HTML]{EFEFEF}44.74 & \cellcolor[HTML]{EFEFEF}46.51 & \cellcolor[HTML]{EFEFEF}27.30 & \cellcolor[HTML]{EFEFEF}64.20 & \cellcolor[HTML]{EFEFEF}47.80 \\
& AMP~\cite{Mugnaini:2025} (IJCNN, 2025) & 56.91 & 52.96 & 52.74 & 33.28 & 67.36 & 52.65 \\
& \cellcolor[HTML]{EFEFEF}\textbf{MoP (Ours)} & \cellcolor[HTML]{EFEFEF}\textbf{61.38$\pm$0.81} & \cellcolor[HTML]{EFEFEF}\textbf{56.27$\pm$1.78} & \cellcolor[HTML]{EFEFEF}\textbf{56.52$\pm$2.99} & \cellcolor[HTML]{EFEFEF}\textbf{34.67$\pm$0.83} & \cellcolor[HTML]{EFEFEF}\textbf{67.85$\pm$1.05} & \cellcolor[HTML]{EFEFEF}\textbf{55.34$\pm$1.42} \\ \hline

\end{tabular}%
}
\label{tab:state_of_the_art}
\end{table*}

\subsection{Experimental Settings}
\label{exp:intro}

\noindent
\textbf{Notable models.} We evaluate MoP on four well-established architectures: the language-only models \changed{LLaMA 7B~\cite{Touvron:2023}, LLaMA-2 7B~\cite{Touvron2:2023}, LLaMA-3 8B~\cite{grattafiori2024llama}, and the multimodal vision-language model LLaVA-1.5 7B~\cite{llava2024improved}}. LLaMA 7B serves exclusively for calibration experiments.

\noindent
\textbf{Evaluation setup.} Building upon previous efforts~\cite{Ma:2023,Kim:2024,Gao:2024disp,Ashkboos:2024,Ouderaa:2024,Mugnaini:2025}, we evaluate the LLMs on five commonsense benchmarks: ARC-e / ARC-c~\cite{Clark:2018}, HellaSwag~\cite{Zellers:2019}, PIQA~\cite{Bisk:2020}, and WinoGrande~\cite{Sakaguchi:2020}. We employ the EleutherAI LM Harness framework~\cite{Gao:2021} to conduct these evaluations. Consistent with established literature~\cite{Ouderaa:2024,Mugnaini:2025}, \changed{we report standard accuracy for WinoGrande and normalized accuracy for the remaining tasks}. For the multimodal architecture LLaVA-1.5\changed{~\cite{llava2024improved}}, we adopt the LMMs-Eval framework~\cite{zhang-etal-2025-lmms} to measure performance on four prominent~\cite{guo2025dynamic,luo2025feast,zhao2024lova,chow2024unified} multimodal datasets: ScienceQA~\cite{lu2022learn}, VizWiz~\cite{gurari2018vizwiz}, LLaVA-Bench (In-the-Wild)~\cite{llava2023}, and MM-Vet~\cite{pmlr-v235-yu24o}. \changed{We report the official metric for each benchmark.}

\noindent
\textbf{Post-training setup.} Adhering to standard practices in the literature~\cite{Ashkboos:2024,YOPO:2025,fu2024amoeballm,Mugnaini:2025,guo2025slimllm,Kim:2024,Gao:2024disp}, \changed{we perform recovery fine-tuning (RFT) on the Alpaca dataset via Low-Rank Adaptation (LoRA)~\cite{Hu:2022}. For the language models, we fix the rank ($r$) at 32 and alpha ($\alpha$) at 10, consistent with previous works~\cite{Gao:2024disp,Ashkboos:2024}}. Furthermore, we employ the AdamW optimizer with a learning rate of 3e-4, a batch size of 16 and train over two epochs, following standard practices~\cite{Ma:2023,Mugnaini:2025}. \changed{For multimodal models, we observe that standard fine-tuning on Alpaca suffices to recover performance (details in Section~\ref{sec:mop-multimodal})}. We therefore retain the LLM configurations, modifying only the rank to 8 and alpha to 16. This adjustment adheres to common literature practices~\cite{Ma:2023,Kim:2024,Mugnaini:2025}. We train both model families using an RTX 4090 GPU. On this hardware, one MoP run to a target compression ratio (including recovery fine-tuning) completes in just 2 hours.

\noindent
\changed{\textbf{Criteria assessment and intermediate fine-tuning.}} We perform all metric evaluations on the WikiText-2~\cite{merity2017pointer} test set, consistent with its standard use as calibration data~\cite{Ashkboos:2024,YOPO:2025,modhifi2025,fu2024amoeballm,Yang:2025Tell,Kim:2024,guo2025slimllm,LinearPatch:2025}. For cosine similarity and Kullback-Leibler divergence, we select 128 random non-empty texts~\cite{Ashkboos:2024,LinearPatch:2025,modhifi2025}. Conversely, perplexity evaluation involves the full test split, processed in fixed segments of 2048 tokens to capture long-range dependencies~\cite{Ashkboos:2024}. As pruning progressively degrades model capabilities, comparing candidates in their raw state yields unreliable predictions of their final post-recovery performance. To mitigate this discrepancy, we apply a dynamic alignment rule: at pruning iteration $i$, we execute $10 \times i$ fine-tuning steps on Alpaca, adhering to the configuration detailed in Section~\ref{exp:intro}. This protocol balances the necessity of restoring minimal model alignment for accurate path selection with the constraint of computational efficiency.

\subsubsection{Comparison of Path Selection Strategies.}

To evaluate the efficacy of the proposed path selection criteria, we compare cosine, KL, PPL, and a random selector at 30\% compression on LLaMA~7B. We report random path as mean and standard deviation over three runs (Table~\ref{tab:ablation}). Across these runs, randomized selection performs on par with PPL and cosine and exceeds KL. Given these results, we adopt random selection as the criterion for the path. This also reduces the computational overhead of the metric calculations. We treat it as a conservative lower bound on MoP’s capability, as MoP is modular and supports stronger future path criteria as drop-in replacements. For all subsequent experiments, we report results over three independent random pruning paths. This finding suggests that the carefully designed matching of pruned parameters in depth and width at each step of MoP is enough to make either path similarly strong, while also substantially improving model performance compared with following only depth or only width, as Figure~\ref{fig:mop_vs_extremes} demonstrates.

\noindent
\subsection{Zero-Shot Performance}

Table~\ref{tab:state_of_the_art} compares MoP against other pruning methods~\footnote{\small{For Shortened LLaMA, we use results from AMP~\cite{Mugnaini:2025}, who ran this method since the original authors do not evaluate this model. For AMP at compression tiers not covered in the original paper, we run the method ourselves. Apart from that, not all methods appear in every comparison, as some original publications do not report results for all models or compression tiers.}}. According to this table, MoP ranks first at 20\%, 30\%, and 40\% compression on both LLaMA-2 7B and LLaMA-3 8B. This tier-wide sweep shifts the interpretation from isolated wins to a robustness guarantee: even the worst MoP seed exceeds the strongest reported baseline mean at the same tier, so MoP does not rely on favorable path selection to achieve state-of-the-art accuracy.

On LLaMA-2 7B, the worst-seed mean accuracy remains 64.97\% at 20\%, 61.24\% at 30\%, and 54.97\% at 40\%. These values exceed the best baseline means at each tier, including AmoebaLLM at 20\% (64.90\%) and AMP at 30\% and 40\% (61.02\% and 53.06\%). The margins become pronounced as compression increases: the worst-seed advantage grows from 0.07 \changed{percentage points (pp)} at 20\% to 1.91 pp at 40\%, and the full MoP mean remains higher still. This pattern indicates that MoP preserves accuracy under progressively tighter capacity constraints instead of trading reliability for compression.

LLaMA-3 8B amplifies the same conclusion. The worst-seed mean accuracy reaches 64.33\% at 20\%, 59.87\% at 30\%, and 54.35\% at 40\%, exceeding the strongest baselines by 1.40 pp at 20\% (CoMe~\cite{Wang:2025CoMe} at 62.93\%), 1.86 pp at 30\% (AMP~\cite{Mugnaini:2025} at 58.01\%), and 1.70 pp at 40\% (AMP~\cite{Mugnaini:2025} at 52.65\%).
\changed{This dominance becomes even sharper when benchmarked against the third-place method at each tier: the margin expands from 1.58~pp at 20\% (LINEARPATCH~\cite{LinearPatch:2025}) to 5.53~pp at 30\% and 6.55~pp at 40\% (both Yang et al.~\cite{Yang:2025Tell}).} Overall, the worst-seed evaluation removes seed selection as an implicit hyperparameter and points out that MoP delivers top-tier accuracy across tiers even under the least favorable path selection. 

\changed{Figure~\ref{fig:mop_vs_extremes} further substantiates the central mechanism behind these gains. When we restrict MoP to prune only in depth or only in width across iterations, the mixed strategy consistently outperforms both single-dimension variants. Collectively, these results confirm that effectively exploiting the complementarity between width and depth establishes a new state-of-the-art}, delivering resilient performance retention even where other strategies falter.

\noindent
\subsection{Inference Speedup}

Practical inference acceleration depends on two critical factors: minimizing the memory bandwidth required to move model weights and shortening the sequential critical path of the forward pass~\cite{Sheng:2023,muralidharan2024compact}. 
MoP optimizes both dimensions simultaneously by removing attention heads, MLP neurons, and entire layers, effectively transforming the original architecture into a leaner dense model that retains computational regularity \cite{Ma:2023,muralidharan2024compact}. 
This explicit structural reduction eliminates the computational overhead of the excised components entirely \cite{Ma:2023,muralidharan2024compact}. 
Such efficiency proves decisive not only for meeting strict latency budgets in real-time applications but also for advancing Green AI goals by materially reducing the energy consumption of large-scale model deployment \cite{Sheng:2023,Schwartz:2020}.

Following Kim et al.~\cite{Kim:2024}, we compute latency on a single NVIDIA RTX 4090 GPU. We define the inference task as processing a prompt of 12 input tokens to autoregressively generate 128 output tokens, using a batch size of 1. To ensure stability, we execute 20 independent runs, discarding the first 10 as warm-up and reporting the mean wall-clock time of the remaining 10. Under this protocol, the dense LLaMA-2 7B model establishes a baseline latency of 2.21s. Relative to this baseline, the pruned models at 20\%, 30\%, and 40\% compression rates achieve latencies of $(1.82 \pm 0.01)$s, $(1.60 \pm 0.03)$s, and $(1.36 \pm 0.03)$s, respectively. Such reductions correspond to speedups of $(1.22 \pm 0.01)\times$, $(1.38 \pm 0.02)\times$, and $(1.63 \pm 0.04)\times$, equivalent to end-to-end latency reductions of $(18.0 \pm 0.7)\%$, $(27.5 \pm 1.1)\%$, and $(38.7 \pm 1.5)\%$, respectively. Comparative analysis reveals that at the 20\% compression tier, MoP outperforms AMP ($1.19\times$) and Shortened LLaMA ($1.13\times$). Furthermore, our method demonstrates superior efficiency against Yang et al.\changed{~\cite{Yang:2025Tell}}: the $1.38\times$ speedup of MoP at 30\% compression distinctively surpasses their reported $1.17\times$ at the same tier and effectively outperforms the $1.30\times$ speedup they achieve at the far more aggressive 50\% compression level.

\noindent
\subsection{MoP on Multimodal Architectures}
\label{sec:mop-multimodal}

\begin{figure}[h]
  \centering
  \includegraphics[width=0.8\linewidth]{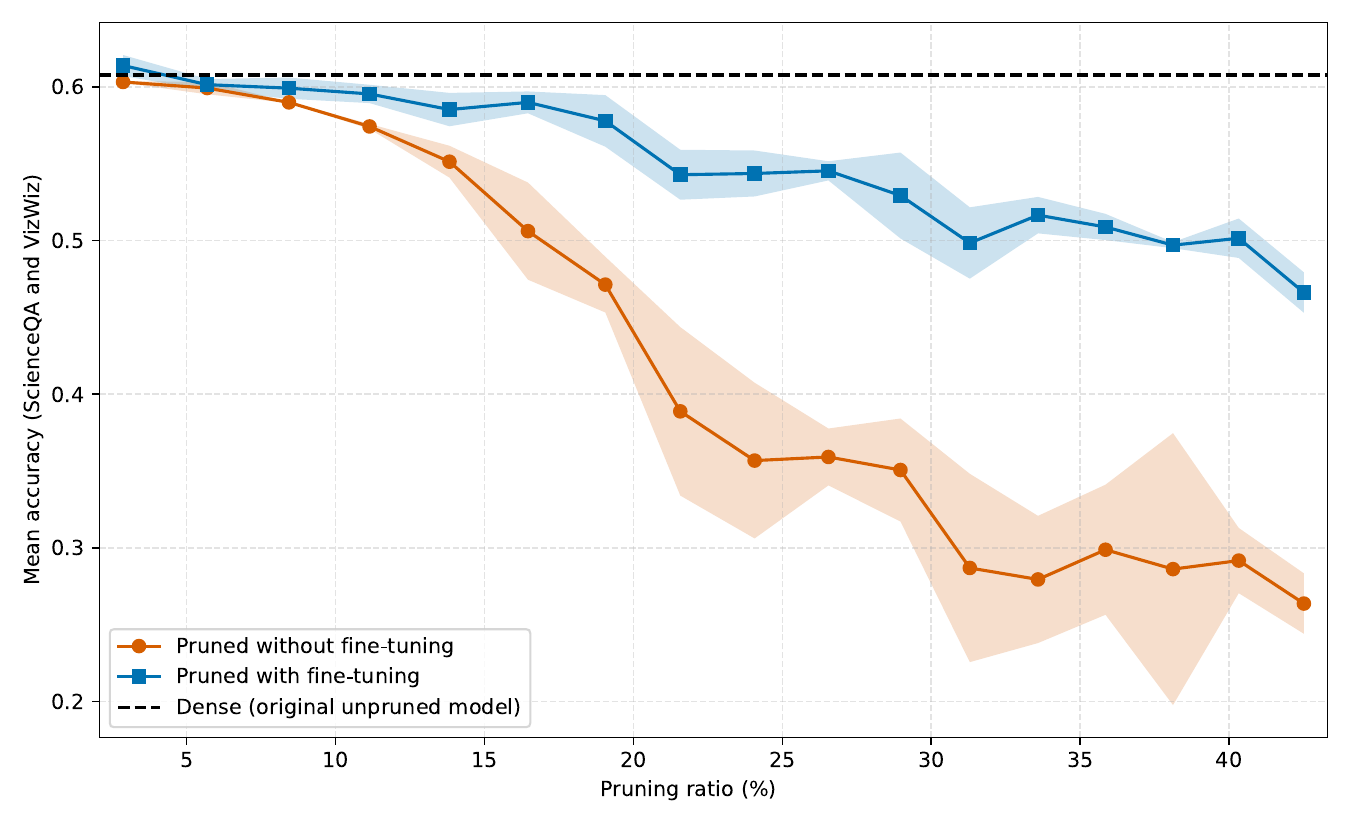}
  \caption{\changed{Impact of pruning on LLaVA-1.5 7B with MoP and recovery with text-only fine-tuning. We present the mean accuracy of three runs for ScienceQA and VizWiz across varying compression ratios. Shaded regions denote the standard deviation.}}
  \label{fig:multimodalgraph}
\end{figure}

Regarding multimodal architectures, we evaluate the performance of MoP on LLaVA-1.5 7B. Figure~\ref{fig:multimodalgraph} illustrates the average accuracy across the ScienceQA and VizWiz datasets for the first 17 iterations of the MoP algorithm. The divergence between the fine-tuned and non-fine-tuned curves emphasizes the necessity of recovery fine-tuning in multimodal pruning and reveals a compelling insight: Alpaca, a dataset designed for text-only models, effectively extends its recovery capabilities to multimodal architectures on visual tasks. To the best of our knowledge, we are the first to document that text-only recovery fine-tuning restores performance in pruned vision–language models. As pruning progresses, the performance gap between the recovered models and their unrefined counterparts becomes pronounced, reaching a peak of 23.71 pp. This disparity extends to stability: while the unrefined models exhibit significant volatility, reaching a standard deviation of 8.85 pp, the fine-tuned variants remain robust, limiting the standard deviation peak to 2.80 pp. Taken together, this experiment underscores the critical role of recovery fine-tuning in the multimodal pruning domain—an area currently under-explored relative to its LLM counterpart—and substantiates the proficiency of Alpaca for RFT despite its text-only origins.

\begin{table}[!b]
\centering
\caption{MoP results on LLaVA-1.5 7B across varying compression rates.}
\renewcommand{\arraystretch}{1.2}
\setlength{\tabcolsep}{8pt} 
\resizebox{\columnwidth}{!}{%
\begin{tabular}{c|c|c|c|c|c}
\hline
Compression & ScienceQA & VizWiz & MM-Vet & LLaVA-Bench & Mean \\
\hline\hline
0\%  & 68.02 & 53.52 & 28.44 & 59.80 & 52.44 \\
\hline
\rowcolor[HTML]{EFEFEF}
20\% & 62.96$\pm$0.33 & 45.60$\pm$3.30 & 23.26$\pm$0.38 & 55.00$\pm$1.35 & 46.70$\pm$0.96 \\
30\% & 61.00$\pm$1.87 & 44.86$\pm$3.93 & 21.94$\pm$1.08 & 50.63$\pm$1.66 & 44.61$\pm$1.98 \\
\rowcolor[HTML]{EFEFEF}
40\% & 57.13$\pm$2.57 & 43.16$\pm$4.78 & 18.88$\pm$1.75 & 44.27$\pm$0.91 & 40.86$\pm$1.22 \\
\hline
\end{tabular}%
}
\label{mult:eval}
\end{table}

Particularly, Table~\ref{mult:eval} details the performance of MoP across three distinct compression regimes: 20\%, 30\%, and 40\%. The analysis expands to include MM-Vet and LLaVA-Bench (In-the-Wild), datasets that introduce significant text generation challenges. Despite this increased complexity, MoP demonstrates significant resilience at 20\% compression, retaining 81.79\% of its original capability on MM-Vet and 91.97\% on LLaVA-Bench (In-the-Wild). At 30\% compression, LLaVA-1.5 exhibits a mean performance drop of 7.83 pp relative to the dense baseline. This decline closely parallels the 7.45 pp drop observed in LLaMA-2 7B at the same compression rate, indicating a consistent degradation pattern across distinct modalities despite the difference in benchmarks. Even at the 40\% compression tier, the model preserves 77.92\% of its mean predictive capabilities. These findings confirm the viability of MoP in diverse compression scenarios and highlight its ability to reduce parameter counts while preserving the core reasoning structures in a multimodal scenario.

\section{Conclusions}\label{sec:conclusions}

Current structured pruning strategies often hit a performance ceiling by restricting the search space to a single axis, either thinning the model width or shortening its depth. MoP overcomes this structural rigidity by unifying both approaches into a cohesive, hybrid strategy. Our results demonstrate that this joint \changed{pruning} yields exceptional resilience: MoP exhibits such stability that its worst-performing path consistently exceeds the average performance of competing methods. This advantage is particularly distinct on LLaMA-3 8B and becomes increasingly pronounced on LLaMA-2 7B as compression intensifies, confirming that the hybrid approach secures robustness exactly where high-compression regimes typically degrade reliability.

Beyond text-only benchmarks, we establish the versatility of MoP in the multimodal domain. By pruning LLaVA-1.5, we effectively preserve performance on visual reasoning tasks, demonstrating a zero-shot cross-modal transferability that purely data-driven, component-specific methods often lack. Furthermore, MoP translates compression into tangible acceleration. The method achieves a $1.38\times$ speedup at 30\% compression, outperforming the $1.30\times$ speedup that Yang et al.\changed{~\cite{Yang:2025Tell}} achieve only at a far more aggressive 50\% compression level. Collectively, these findings validate the hybrid width-depth paradigm as the superior standard for efficient LLMs and a promising avenue for the multimodal domain.

\section{Acknowledgments}
The authors would like to thank Instituto de Ciência e Tecnologia Itaú (ICTi) and Programa de Bolsas Itaú (PBI). This study was financed, in part, by the São Paulo Research Foundation (FAPESP), Brasil. Process Number \#2023/11163-0. This study was financed in part by the Coordenação de Aperfeiçoamento de Pessoal de Nível Superior – Brasil (CAPES) – Finance Code 001. The authors would like to thank grant \#402734/2023-8, National Council for Scientific and Technological Development (CNPq). Anna H. Reali Costa would like to thank grant \#312360/2023-1 CNPq. 
This work was partially supported by the Instituto Nacional de Ciência e Tecnologia em Inteligência Artificial Responsável para Linguística Computacional, Tratamento e Disseminação de Informação (INCT-TILDIAR - CNPq grant \#408490/2024-1).

\bibliography{refs}
\bibliographystyle{icml2026}

\end{document}